\documentclass[final,5p,times,twocolumn,authoryear]{elsarticle}

\usepackage{amssymb}
\usepackage{lipsum}
\usepackage{amsmath}

\journal{Elsevier}
\usepackage{lineno}
\begin{document}
\begin{frontmatter}


\title{YOLO11 and Vision Transformers based 3D Pose Estimation of Immature Green Fruits in Commercial Apple Orchards for Robotic Thinning}

\author[1]{Ranjan Sapkota\corref{cor1}}
\author[2]{Manoj Karkee}

\affiliation[1]{organization={Biological \& Environmental Engineering}, 
            addressline={Cornell University}, 
            city={Ithaca}, 
            postcode={14850}, 
            state={NY}, 
            country={USA}}

\affiliation[2]{organization={Center for Precision and Automated Agricultural Systems, Department of Biological Systems Engineering}, 
            addressline={Washington State University},  
            state={WA}, 
            country={USA}}

\cortext[cor1]{Corresponding author}
\ead{rs2672@cornell.edu}

\begin{abstract}
Manual thinning of immature green apples in commercial orchards is labor-intensive and costly, necessitating the development of automated solutions to address labor shortages and improve efficiency. This study developed a robust method for 3D pose estimation of immature green apples (fruitlets) in commercial orchards using YOLO11 object detection model and Vision Transformers for depth estimation.  The results revealed that YOLO11n outperformed other configurations in box and pose precision (0.91 and 0.92, respectively) and demonstrated the fastest inference speed of 2.7 ms per image. YOLO11s achieved the highest box mAP@50 score of 0.94, while YOLOv8n reached the highest pose mAP@50 score of 0.96. For depth estimation, Depth Anything V2 showed superior performance with the lowest RMSE of 1.52 and MAE of 1.28 in 3D pose length validation. These findings highlight the potential of integrating YOLO11 and Vision Transformers, particularly Depth Anything V2, for accurate and efficient 3D pose estimation of immature green fruits, paving the way for future robotic thinning applications in commercial orchards, which is expected to reduce the labor demands in orchard operations.
\end{abstract}




\end{frontmatter}




\section{Introduction}
\label{introduction}
Labor shortages continue to present significant challenges in U.S. agriculture, prompting the need for advancing automation and robotics solutions to minimize the dependence on manual labor and enhance crop production \cite{richards2024labor, karan2021resilience}. Technologies such as precision planting systems automate seed placement with exact spacing and depth \cite{khadatkar2021development}, optimizing growth conditions and reducing the need for manual labor. Similarly, robotic weed control systems commercialized in recent years use advanced sensors and precise application/weeding techniques to identify and eliminate weeds without human intervention \cite{wu2020robotic, upadhyay2024advances}. Researchers and technology companies are also developing fruit-picking robots with capabilities to harvest crops such as apples and citrus as fast or faster than human pickers and with greater accuracy  \cite{xiao2024review, fei2021co}. 

Despite advancements in automation and robotics for various types of crops, field operations in  tree fruit crops such as apples (Malus domestic), peaches (Prunus persica) and cherries (Prunus avium) remain labor-intensive \cite{kaur2023insights, rutledge2023farm}. For example, Cherries require careful hand-picking due to their delicate nature, to prevent damage to both the fruit and the trees—a task that current machinery cannot perform with the necessary gentleness \cite{karkee2018mechanical, zhou2016effect}. Apples and many other fruit crops are also harvested manually despite recent commercialization efforts \cite{musacchi2018apple}. Plums (Prunus domestica) and apples need selective pruning to maintain desired canopy structure that is essential for improved fruit quality and manage tree health \cite{seehuber2011regulation}. Furthermore, these crops depend heavily on manual thinning to optimally distribute fruit across canopy areas, which is essential to optimize fruit size and quality, and to minimize biennial bearing \cite{iglesias2023peach, costa2022thinning}. These manual operations create a substantial reliance on decreasing human labor, highlighting the need for innovative labor-saving solutions. 

\begin{figure*}[ht]
\centering
\includegraphics[width=0.85\linewidth]{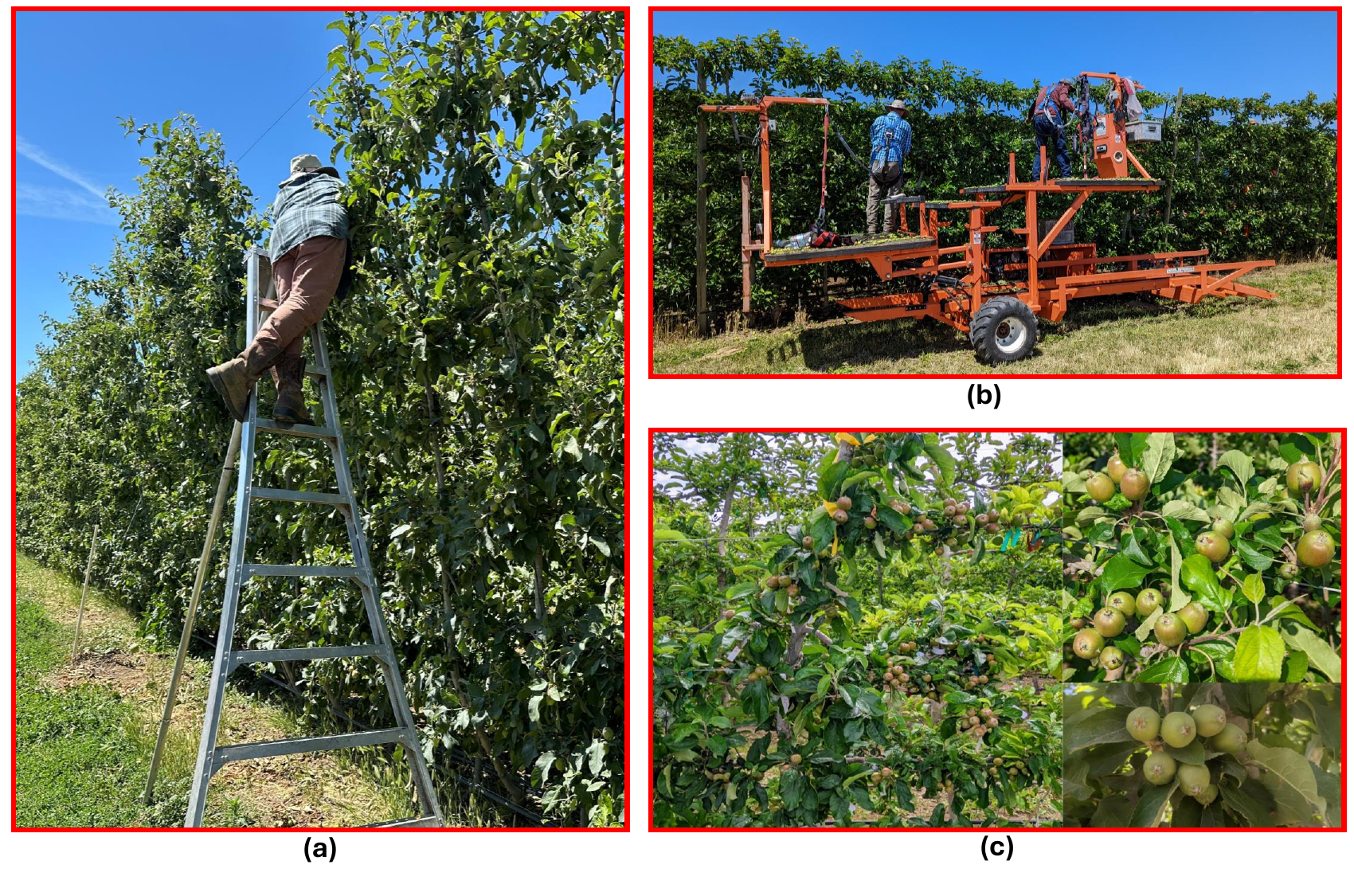}
\caption{Manual thinning of apple fruitlet in a commercial 'Scilate' apple orchard in Prosser, Washington State, USA, during 2024 growing season; (a)  use of an aluminum ladder by farm workers to access fruit clusters in upper canopy region, demonstrating the physical demands and safety risks of traditional methods; b) manual thinning using a mechanized platform, which reduced the need for climbing ladders, but still required extensive manual efforts to reach fruitlet clusters; c)  Showing the high-density examples of immature green apples in commercial orchards which needs thinning.}
\label{fig:Figure1}
\end{figure*}

Among these crop-load management operations, immature green fruit thinning is one of the crucial operations used to optimize fruit size and quality by reducing competition for water, sunlight, and nutrients among fruits \cite{dennis2000history, costa2018fruit}. This practice is particularly crucial for apple production, an important crop in the United States \cite{statista-2022} and globally  \cite{o2021economic}. The labor-intensive nature of manual thinning (\ref{fig:Figure1}), combined with persistent labor shortages and the increasing costs associated with securing skilled labor, highlight the urgent need for innovative solutions to automate this process \cite{assirelli2018evaluation, lordan2018screening}. Each year, over a hundred thousand workers come to the US for farm labor \cite{holmes2023fresh, weiler2022seeing}. Figure \ref{fig:Figure1} shows how these workers are engaged in physically demanding agricultural tasks such as green fruitlet thinning in commercial orchards. 

As shown in \ref{fig:Figure1}a, workers often move between trees with carrying and setting an aluminum ladder, and then perform these farming tasks while claiming up and down the ladder that are often 14 feet or higher. This method not only demands physical agility to climb and reposition the ladder continuously but also poses safety risks inherent in working at heights and with repeated hand movements.

Figure \ref{fig:Figure1}b depicts an alternative approach where multiple workers are using a mechanized platform that moves through orchard rows (either automated or driven by an operator). While this method reduces the need to climb and improves the working condition compared to ladder-based operation, it still requires significant manual effort. Workers must bend and stretch extensively to reach fruitlets in dense clusters, as shown in Figure \ref{fig:Figure1}c. This physically demanding work not only entails high labor costs but also poses significant health risks to workers, as mentioned before, particularly related to spinal injuries and other musculoskeletal disorders. Studies have highlighted that the repetitive and strenuous postures required in manual thinning, such as excessive back bending and upper arm elevation, can lead to severe physical strain \cite{thamsuwan2020comparisons}. Specifically, while platforms can reduce upper arm strain compared to ladders, they do not significantly alleviate the stress on workers' backs, who must still engage in substantial torso flexion. Furthermore, the agricultural sector is notorious for the prevalence of upper limb injuries, with a systematic review revealing that a substantial number of these injuries occur during manual tasks like harvesting and thinning, often leading to long-term disabilities and economic consequences \cite{mucci2020upper}.

Despite some advancements in mechanization, manual labor remains prevalent in fruitlet thinning due to the precision required for tasks like immature apple detection, where occlusion by leaves, overlapping fruits, and variable lighting conditions present significant challenges \cite{linker2012determination, sun2019recognition}. Studies have explored deep learning approaches for detecting young apples with some success, yet issues with accuracy and processing speed persist \cite{xia2018detection, tian2019apple, huang2021immature}. Techniques like Lidar-based modeling for 3D pose estimation of green pepper where the orientation of a fruit in space is obtained by fitting a model to surface points of the fruit \cite{eizentals20163d}, low-cost RGB-D sensors with multiple three-dimensional (3D) line-segments detection   \cite{lin2019guava},  and binocular imagery \cite{yin2021fruit} have shown high precision and recall in pose estimation for various fruits. However, these methods often struggle with real-time processing requirements and complex environments, making them less feasible for field applications. Therefore, there is a need for faster, more precise, and simpler fruit thinning systems capable of operating effectively in complex and dynamic orchard environments. While innovative approaches including the use of single RGB images for citrus pose estimation \cite{sun2023citrus} and detection of symmetry axes in 3D point clouds for in citrus pose estimation from an RGB image for automated harvestingin, and  sweet peppers pose estimation\cite{li2018pose} are promising, their dependence on specific conditions.

To address these challenges, this study focused on developing a machine vision system for 3D pose estimation of immature green fruits in commercial apple orchards, which will provide a foundation for the development of robotic fruit thinning technologies. The specific objectives are as follows
\begin{itemize} 
\item \textbf{Data Acquisition:} Deploy a machine vision system on a robotic platform to collect high-resolution RGB images within a commercial orchard, creating a comprehensive dataset for model training and testing. 
\item \textbf{Deep Learning-based Detection and Pose Estimation:} Utilize the YOLO11 model to accurately detect and determine the position and orientation of immature green fruits, enhancing the precision with uniform hyperparameter settings across the models. 
\item \textbf{RGB to RGB-D Mapping Using Vision Transformers:} Convert RGB images into 3D point clouds using Vision Transformers, specifically the Dense Prediction Transformer (DPT) and Depth Anything V2, to improve pose estimations.
\item \textbf{Field Level Validation}: Validate these estimations against field-level measurements of the major axis values of immature green apple fruits, ensuring the accuracy and reliability of the vision system for practical applications. 
\end{itemize}

\section{Related Work}
\subsection{Deep Learning-based Fruit Detection and Pose Estimation:}
Traditional methods for apple and other fruit detection have mostly focused on mature fruit suitable for robotic harvesting and crop-load estimation, utilizing basic image processing as well as deep learning techniques. These methods, however, often face challenges in detecting immature apples (desired for robotic fruitlet thinning) in complex natural environments where young fruits exhibit colors and textures similar to foliage \cite{kataoka1998automatic, xuan2020apple}. To address this challenge, Xia et al. (2018) \cite{xia2018detection} employed an innovative method combining adaptive green and blue chromatic aberration mapping with an iterative threshold segmentation algorithm, achieving a true positive rate of 88.51\% and an F1-Measure of 90.29\% in detecting young green fruit, which was a notable improvement compared to past studies in outdoor orchard environments. 

Recent advancements in machine learning and deep learning have introduced more sophisticated models, such as YOLO series of models, which significantly enhance detection capabilities by learning from diverse datasets without explicit programming \cite{liakos2018machine, xuan2020apple}. Utilizing these advancements, a few studies have focused on detecting immature green apples in complex orchard environments, which is crucial for automating the thinning process. For example, Tian et al. (2019) \cite{tian2019apple} refined the YOLO-V3 model using DenseNet processing, enhancing apple detection capabilities across various growth stages under challenging conditions such as fluctuating illumination and fruit overlap, demonstrating the potential for real-time applications. Additionally, Huang et al. (2021) \cite{huang2021immature} further advanced the YOLOv3 framework for identifying immature apples, achieving high-speed detection of 83 frames per second on a 1080ti GPU, effectively addressing occlusions and overlapping fruits. Despite these efforts, there is still a lack of studies targeting the development of comprehensive vision models capable of accurately estimating fruit location, orientation, and size, which is essential for developing robotic systems of immature green apple thinning. This task is particularly challenging because of the early growth stage of fruit, when issues such as occlusion and overlap of fruits by leaves, branches, and other fruit, as well as variable lighting conditions, are more prominent \cite{linker2012determination, sun2019recognition}.

It is also noted that there has been a wide range of studies on other fruit crops showing considerable advancements in fruit detection and pose estimation, albeit with shared limitations. Eizentals et al. (2016) \cite{eizentals20163d} utilized Lidar-based model matching for 3D pose estimation of green peppers, achieving effective results under controlled conditions but struggling in natural, cluttered settings. Lin et al. (2019) \cite{lin2019guava} applied a low-cost RGB-D sensor for guava detection, achieving high accuracy but facing challenges in real-time processing, highlighting the need for faster methods. Similarly, Yin et al. (2021) \cite{yin2021fruit} used binocular imagery and neural networks for detecting grape clusters, showing high accuracy but requiring complex setups that hinder practical field application. These studies underscore the potential of advanced imaging and machine learning in agricultural applications while also highlighting critical areas for improvement, particularly in processing speed and operational simplicity. 

Studies in pose estimation have also extended to crops like tomatoes and peppers, with researchers addressing unique challenges in these domains. Kim et al. \cite{kim20232d} developed a 2D pose estimation method that effectively identified tomato-pedicel pairs, though its reliance on visible keypoints limited its effectiveness under occlusion. Li et al. (2018) \cite{li2018pose} introduced a pose estimation strategy for sweet peppers using symmetry axes in point clouds, which varied in performance based on the physical state of the peduncle, introducing inconsistencies in uncontrolled environments. While the studies based on deep learning methods in fruit detection and pose estimation are expanding, the common challenges across the studies include the need for higher accuracy, faster processing speeds, and simpler, more robust systems capable of functioning effectively in dynamically changing and visually complex orchard environments.

\subsection{Vision Transformers Application in Agriculture}
Vision Transformers (ViTs), which utilize the self-attention mechanism to process images as sequences of patches, have emerged as powerful alternatives to convolutional neural networks for various image analysis tasks \cite{thakur2021vision, li2022plant} . ViTs have increasingly been recognized for their powerful capabilities in addressing complex challenges in agriculture, particularly in the realm of plant disease classification. Studies such as that by Salamai et al. (2023) \cite{salamai2023lesion} have showcased the effectiveness of ViTs in detecting paddy diseases, where the lesion-aware visual transformer model achieved an impressive accuracy of approximately 98.74\% and an F1-score of around 98.18\%. This performance highlights the potential of ViTs in precision agriculture for enhancing disease identification with high accuracy and efficiency. Similarly, Thai et al. (2023) \cite{thai2023formerleaf} introduced FormerLeaf, a transformer-based model that incorporates the Least Important Attention Pruning (LeIAP) technique and sparse matrix-matrix multiplication (SPMM) to optimize the detection of leaf diseases. This model not only reduced training times but also significantly improved accuracy, demonstrating the adaptability of ViTs in agricultural settings. Furthermore, Thakur et al. (2021) \cite{thakur2021vision} combined conventional Convolutional Neural Networks (CNNs) with Vision Transformers to create PlantViT, a hybrid model for diagnosing plant diseases from leaf images, achieving notable accuracy across extensive datasets. This hybrid approach underscores the robustness and generalizability of ViTs in diverse agricultural environments.

In addition to disease detection, ViTs have also been applied to other agricultural tasks such as crop yield prediction and multi-object tracking, illustrating their versatility. Lin et al. (2023) \cite{lin2023mmst} developed a Multi-Modal Spatial-Temporal Vision Transformer (MMST-ViT) model, which integrates visual remote sensing data with meteorological inputs to evaluate the influence of environmental factors on crop yields. This model utilizes a novel multi-modal contrastive learning technique for effective pre-training without extensive human intervention, significantly outperforming existing models in several key performance metrics. Specifically, MMST-ViT excelled in metrics such as Root Mean Square Error (RMSE), R-squared, and Pearson Correlation Coefficient, underscoring its superior predictive accuracy in agricultural settings. Additionally, Hernandez et al. (2024) \cite{hernandez2024multi} proposed a Multi-Object Tracking (MOT) framework using a Vision Transformer, tailored to agricultural environments. This framework employed the local feature matching transformer (LoFTR) to improve spatial associations among plants within cluttered fields, significantly enhancing the robustness and accuracy of robotic tracking applications. Tested on specialized datasets such as LettuceMOT and AppleMOT, this method achieved up to a 44\% improvement in tracking accuracy over existing solutions \cite{hernandez2024multi}. These advancements demonstrated the potential of ViTs to manage spatial dynamics effectively, facilitating the adoption and advancement of robotic technologies in agriculture.

While Vision Transformers have not yet been widely applied in automating orchard operations, their demonstrated success in various other agricultural applications shows the potential for exploring their applications to this area. Specifically, based on their robust performance in disease classification, crop yield prediction, and object tracking, there is a promising opportunity to utilize ViTs for enhancing object detection, segmentation, localization, and tracking in fruit tree canopies, thus helping improve automation and robotics technologies for orchard operations.

\section{Methods}
\begin{figure*}[htbp]
\centering
\includegraphics[width=\linewidth]{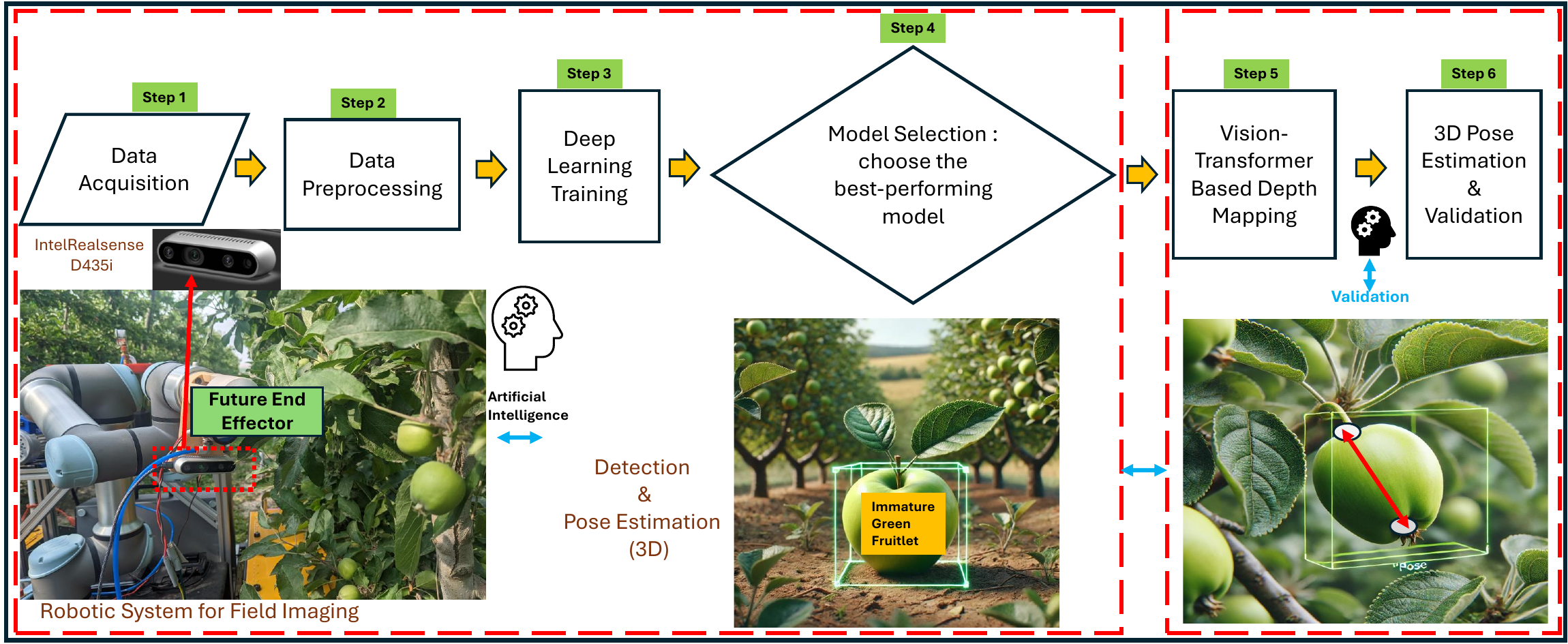}
\caption{Outlining of the research the methodology that includes various steps including data collection (using a robotic platform), deep learning model training and validation, and precise pose estimation of green fruitlets.}
\label{fig:Method1}
\end{figure*}

The overall methodology of this study is summarized in Figure \ref{fig:Method1}. As show in the figure, the workflow consists of data collection, pre-processing, and the deployment of deep learning models for pose estimation. A dataset consisting of RGB and Depth images was collected from a commercial apple orchard using a robotic platform. The images obtained were manually annotated, which involved precise delineation of bounding boxes and estimation of pose lines on immature green fruits, crucial for the subsequent model training phase. These annotated datasets were employed to train various configurations of the YOLO11 and YOLOv8 models for comparative assessment of their performances. The goal of this step was to identify the most effective model configuration for fruitlet pose estimation based on the precision and inference speed achieved. The optimal model identified from this comparative assessment was then used to transform RGB images into RGB-D point clouds. This process to generate 3D point clouds utilized monocular images and two state-of-the-art Vision Transformers: Depth Anything and Dense Prediction Transformer. These tools mapped the RGB properties of the images to simulate depth, creating depth-enhanced RGB-D data necessary for accurate 3D representation and enhanced pose estimation. The conversion from RGB to RGB-D data was also critical for the validation process as depicted in the right side of Figure \ref{fig:Method1}. The poses estimated by the YOLO11 model were validated against actual field measurements obtained using a digital caliper. This validation process involved estimating the major axis length of each fruitlet in the RGB-D point clouds and comparing these measurements with the corresponding field measurements to ensure the accuracy and reliability of the pose estimation models. In the following several sub-sections, these steps of the research methodology, including "Data Acquisition", "Model Training" and "Field Validation" will be described in more details.

\subsection{Data Acquisition and Preparation}
Data acquisition was carried out using a robotic platform that was equipped with an Intel RGB-D camera, mounted on a UR5e robotic arm as depicted in Figure \ref{fig:Method1}. This setup facilitated the capture of RGB images of immature fruitlets in a commercial "Scifresh" apple orchard located in Prosser, Washington State, USA. The images were collected prior to fruitlet thinning in May 2024, with the appropriate timing for fruitlet thinning determined through regular monitoring of the commercial apple orchard and ongoing communication with growers and farm workers. The orchard used was planted in 2008 with tree row spacing of 3 meter and an intra-row tree spacing of 1 meter.

In this study, a total of 1,147 images were collected using Intel RealSense D435i camera. This camera features a depth-sensing system utilizing active infrared (IR) stereo vision and an inertial measurement unit (IMU). The camera’s depth sensor operates using structured light technology, which employs a pattern projector to create disparities between stereo images captured by two IR cameras. The 3D sensor of this camera has a resolution of 1280 × 720 pixels and can capture depth information up to a range of 10 m. The camera supports a frame rate up to 90 frames per second (fps), and has a horizontal field of view (HFOV) of 69.4° and vertical field of view (VFOV) of 42.5°. Additionally, the 6-axis IMU provides precise orientation data, enhancing the alignment of depth data and scene understanding.

\begin{figure*}[ht]
\centering
\includegraphics[width=\linewidth]{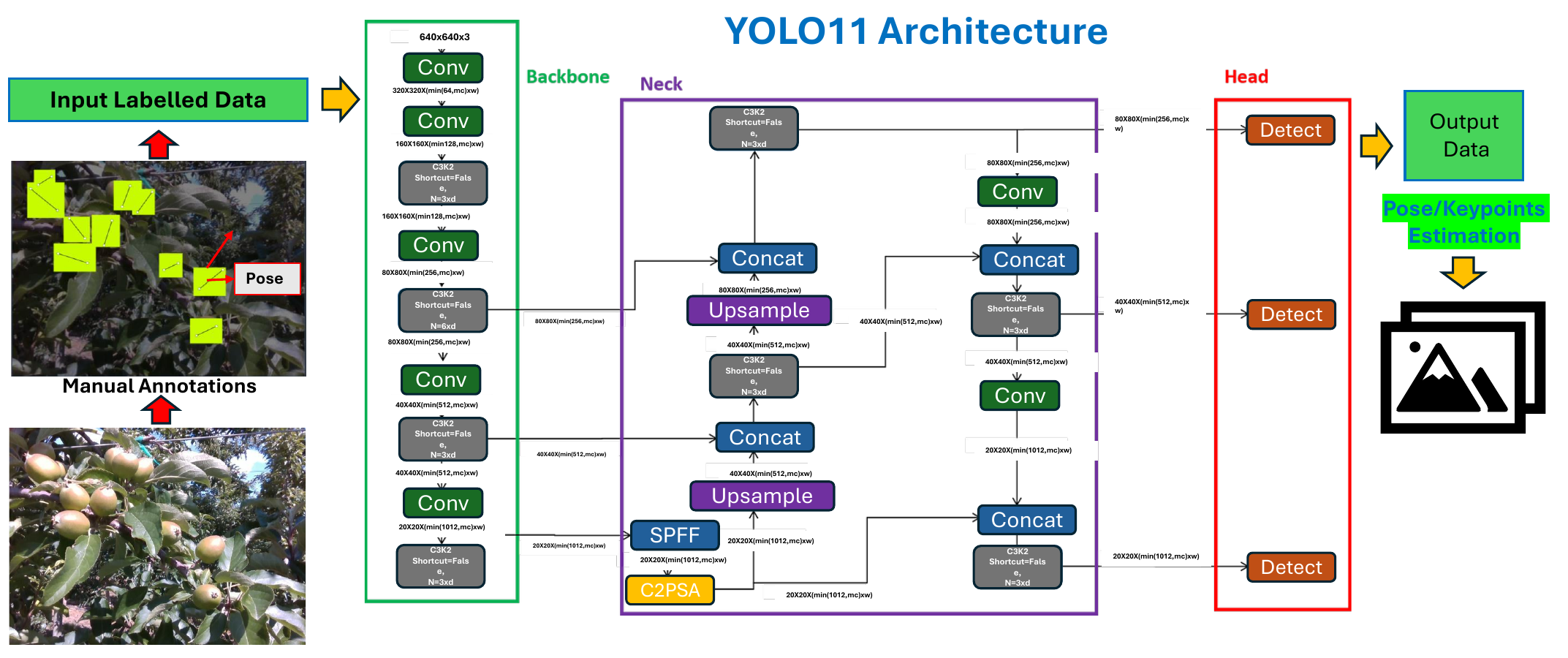}
\caption{YOLO11 Architecture used for immature green fruit detection}
\label{fig:YOLO11architecture}
\end{figure*}

The pre-processed images manually annotated in roboflow (Roboflow, Des Moines, IOWA) manually  to create the dataset for model training and testing. The annotation process included identification of y crucial keypoints on immature fruitlets, specifically the calyx and peduncle, as illustrated in Figures \ref{fig:Method1} and \ref{fig:YOLO11architecture}. A bounding box and a line were then used to denote the object and its pose, respectively. This was a labor-intensive annotation process, which was vital for generating accurate object labels essential for training deep learning models. Annotated dataset was then divided into 80\% (918 images) for training, 10\% (114 images) for validation, and 10\% (115) for testing. This organized dataset was then exported using Roboflow, ensuring structured access for model development and training.

\subsection{YOLO11 Model Overview and Training for Pose Estimation}
In this study, we employed the YOLO11 (Glenn Jocher 2024 \cite{yolo11_ultralytics}) model for its advanced capabilities in detecting and estimating the pose of immature green fruits in orchard environments. YOLO11 (Figure \ref{fig:YOLO11architecture}) represents the latest advancement in the YOLO series, building on the strengths of its predecessors with innovative features and optimizations that enhance accuracy and processing speed, making it ideal for real-time applications in agricultural environments where rapid and precise object detection is crucial. Specifically, the YOLO11 architecture has optimized feature extraction capabilities as depicted in Figure \ref{fig:YOLO11architecture}, enabling it to capture fine details in images. This model supports a wide range of applications, including real-time object detection, instance segmentation, and pose estimation. It is capable of accurately identifying objects irrespective of their orientation, scale, or size, making it highly versatile for use in industries such as agriculture and surveillance. YOLO11 incorporates advanced training techniques that have led to significant improvements on benchmark datasets. For example, the YOLO11m variant achieved a higher mean Average Precision (mAP) on the COCO dataset while utilizing 22\% fewer parameters than YOLOv8m, showcasing enhanced efficiency without sacrificing accuracy. Designed with an inference speed that surpasses earlier models, YOLO11 is optimized for real-time applications, ensuring fast processing even in demanding environments.

The training process for the YOLO11 models was structured to optimize performance across various computer vision tasks. The model was trained with a batch size of 8 and a uniform image resolution of 640x640 pixels. The training process utilized an automatic optimizer and implemented a total of 100 epochs, with a patience setting of 100 to avoid over-fitting. Key hyperparameters optimzied included an initial learning rate of 0.01, momentum at 0.937, and weight decay set at 0.0005. A warmup phase of 3 epochs was integrated to stabilize training parameters, alongside specific loss settings including box loss of 7.5 and class loss of 0.5. The default model image augmentation techniques used were flipping, translation, and a unique mosaic approach with a probability of 1.0. Moreover, the model was designed to operate with advanced features like overlap masks and a dynamic workspace of 4. In this work, pre-initialized weights were not used, ensuring that the models were trained to adapt solely based on the dataset collected in this study, thus providing a fresh foundation to assess the effectiveness of the training and data augmentation strategies employed. 

To meet the significant computational needs of this study, the training was conducted on a high-performance computing system. This system featured an Intel Xeon(R) W-2155 CPU with a base clock speed of 3.30 GHz and 20 cores. For graphics processing and machine learning tasks, the system was equipped with NVIDIA Corporation GP102 [TITAN Xp] graphics cards, adept at handling complex image processing tasks. The workstation also boasted a large storage capacity of 7.0 TB, facilitating extensive data management and analysis. It ran on Ubuntu 20.04.6 LTS, a robust and stable 64-bit operating system. The graphical interface of the system was managed by GNOME version 3.36.8, and X11 was used as its windowing system, creating a stable and efficient environment for the computational analyses required to train the YOLO11 configurations.  

After the YOLO11 training was completed, its predecessor, YOLOv8, along with its five configurations (developed by the same company, Ultralytics), were thoroughly evaluated. Each configuration of YOLOv8 was tested under identical hyperparameter settings and using the same workstation to ensure a controlled and fair comparison.

\subsection{ Performance Evaluation for Detection and Pose Estimation}
The performance of YOLO11 model in detecting immature green apples and estimating their pose was evaluated using \textit{precision}, \textit{recall}, and \textit{mean Average Precision} at 50\% Intersection over Union (mAP@50). These metrics are further defined as follows:

\begin{itemize}
    \item \textbf{True Positives (TP):} The number of immature green fruits correctly identified by the model.
    \item \textbf{False Positives (FP):} The number of instances where the model incorrectly identifies an object as an immature green fruit.
    \item \textbf{True Negatives (TN):} The number of instances correctly identified as not being immature green fruits. (Note: This term is generally not used in the calculation of precision and recall for object detection tasks.)
    \item \textbf{False Negatives (FN):} The number of immature green fruits present in the image that the model fails to detect.
\end{itemize}

\textbf{Precision} for both box detection and pose estimation is defined as the ratio of correctly predicted positive observations to the total predicted positives (Equation \ref{eq:precision}):
\begin{equation}
\text{Precision} = \frac{TP}{TP + FP}
\label{eq:precision}
\end{equation}

\textbf{Recall}, shown in Equation \ref{eq:recall}, measures the model's ability to detect all relevant instances:
\begin{equation}
\text{Recall} = \frac{TP}{TP + FN}
\label{eq:recall}
\end{equation}

\textbf{Mean Average Precision at 50\% Intersection over Union (mAP@50)} considers the intersection over union (IoU) between the predicted box and the ground truth box, where a detection is considered correct if the IoU is greater than 50\%. It is averaged over all classes and IoU thresholds (Equation \ref{eq:map50}):
\begin{equation}
\text{mAP@50} = \frac{1}{N} \sum_{i=1}^{N} \left( \frac{TP_i}{TP_i + FP_i + FN_i} \right)
\label{eq:map50}
\end{equation}

These metrics (\ref{eq:precision}, \ref{eq:recall}, and \ref{eq:map50}) provide a comprehensive view of the model's performance in detecting and estimating the pose of immature green fruits, highlighting the effectiveness and accuracy of the models used.

Additionally, the evaluation of image processing speed for green fruit pose detection was systematically divided into three phases: pre-processing, inference, and post-processing. These phases encompass the complete cycle from initial image manipulation before detection, through the detection process itself, to the final steps following the detection. Each phase is critical in understanding the overall efficiency and responsiveness of the pose detection system.

\begin{itemize}
    \item \textbf{Pre-processing:} This stage involves preparing the images for detection, which may include resizing, normalization, and augmentation to enhance the model's ability to detect immature green fruits accurately.
    \item \textbf{Inference:} At this core phase, the model processes image to detect fruit and estimate their pose. The speed of this stage is crucial for real-time applications and is quantified by the time taken to process a single image, which is given by \ref{eq:inference_speed}:
    \begin{equation}
    \text{Inference Speed} = \frac{\text{Total Processing Time}}{\text{Number of Images Processed}}
    \label{eq:inference_speed}
    \end{equation}
    \item \textbf{Post-processing:} This phase involves operations after the initial detection, such as applying non-maximum suppression to refine the detection boxes and extracting pose information for further analysis.
\end{itemize}
Based on achieving the highest precision and the fastest inference speed among all configurations of YOLO11 and YOLOv8, the optimal model was selected for further image analysis and validation of the detected pose in 3D image space.

Based on these measures, the best-performing YOLO11 configuration was selected from all tested variants of YOLO11 and YOLOv8 pose detection models. The 2D Pose estimation of immature green apples from the selected  model configuration were used to implement two Vision Transformer models; DPT and Depth Anything V2. These models transformed standard RGB images into enriched RGB-D point clouds, essential for advancing detailed 3D mapping capabilities. This step was pivotal in integrating precise depth information with RGB image, which is expected to enhance the robotic technologies for effective thinning operations in commercial apple orchards.

\subsection{Vision Transformers for RGB to RGB-D Mapping}
The two Vision Transformers, DPT and Depth Anything V2, are explored in this section for point cloud generation based on RGB image. Overall block-diagram for this step is presented in Figure \ref{fig:bothVits}a whereas detailed workflow for each model is illustrated respectively in Figures \ref{fig:bothVits}a, and \ref{fig:bothVits}b. 

DPT, designed primarily for depth estimation, utilizes a transformer architecture that is effective in processing high-resolution RGB images to generate depth-enhanced RGB-D data. This model leverages pre-trained weights and a sophisticated feature extraction process to estimate precise depth information, which is essential for accurate 3D mapping of objects. This model uses a lightweight transformer model that supports faster computation while maintaining high accuracy. The model is particularly effective in generating detailed depth maps from RGB images, which are essential in calculating the dimensions and orientations of objects (e.g., size, volume, orientation , distance, texture, segmentation, and camera perspectiveetc) needed for robotic thinning operations in orchards.

\begin{figure*}[ht]
\centering
\includegraphics[width=0.7\linewidth]{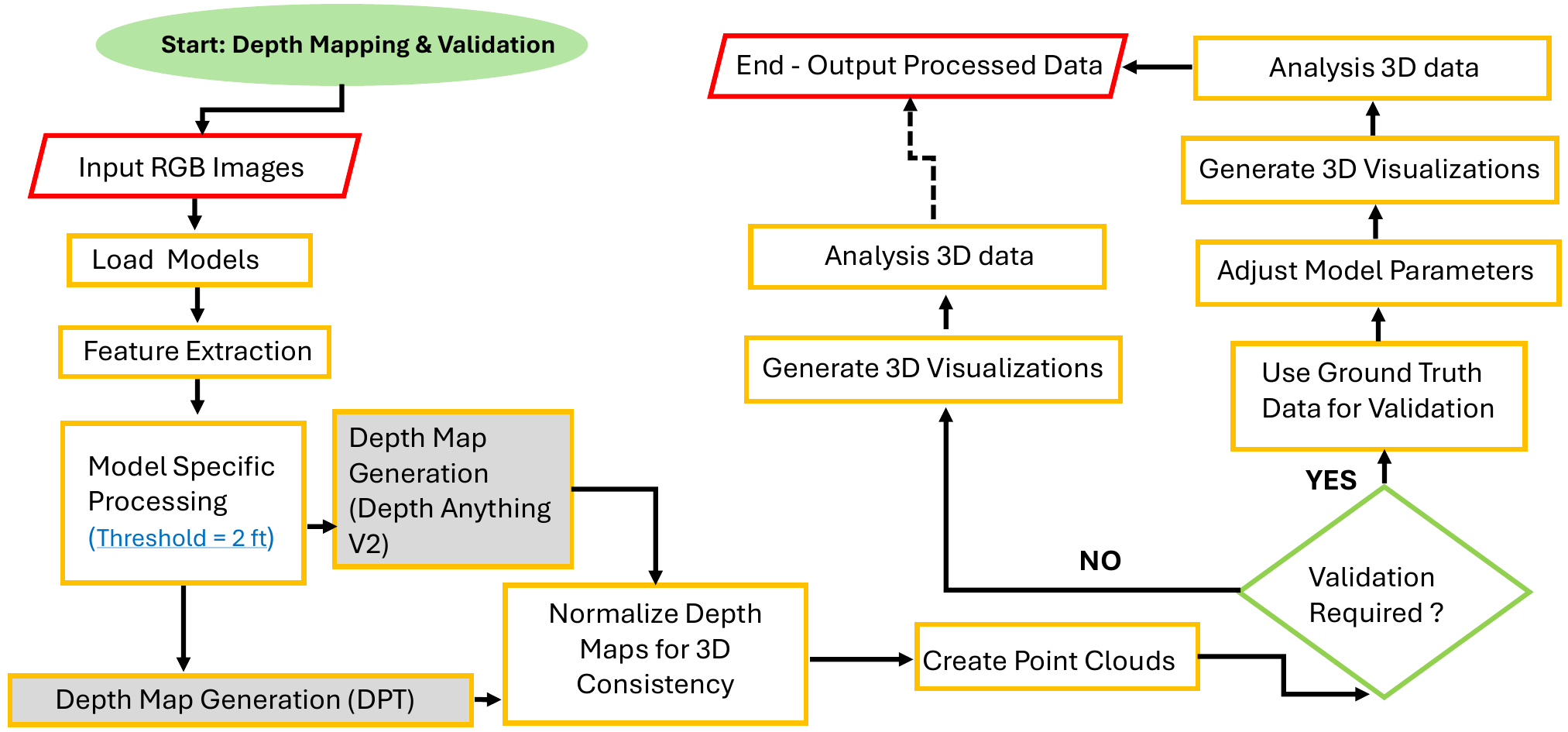}
\caption{Flowchart of the methodology used for depth mapping and validation using two Vision Transformer models. RGB images of apple trees with  immature green apples (fruitlets), initially processed using YOLO11n for pose estimation, were transformed into RGB-D data by Dense Prediction Transformer (DPT) and Depth Anything V2. The depth maps generated were then used to create precise 3D point clouds, which were validated against ground truth measurements.}
\label{fig:TransformersBlockFlow}
\end{figure*}
\begin{figure*}[ht]
\centering
\includegraphics[width= 0.9\linewidth]{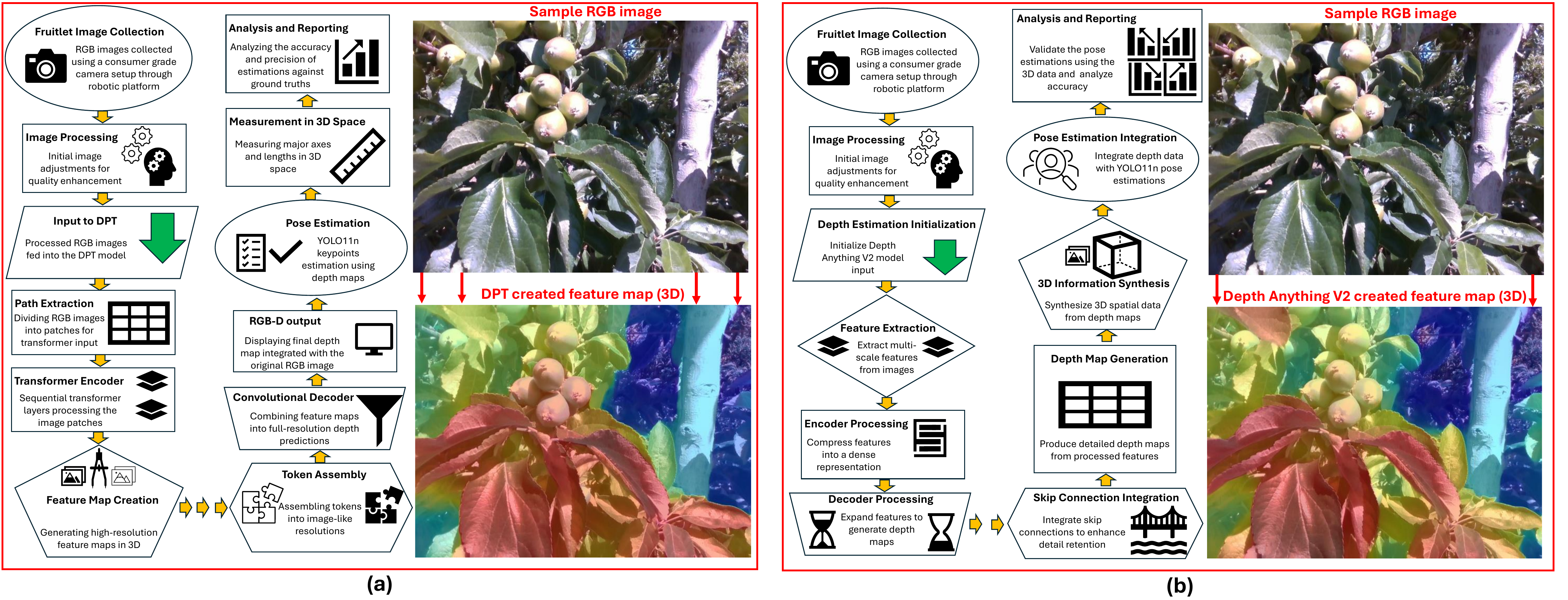}
\caption{ (a) Workflow of the Dense Prediction Transformer (DPT) for generating 3D point clouds. DPT processes RGB images, originally captured using a consumer-grade camera, to produce RGB-D data, facilitating the creation of detailed 3D point clouds; (b) Workflow of Depth Anything V2, detailing each step from RGB image processing to depth map generation and subsequent 3D point cloud generation}
\label{fig:bothVits}
\end{figure*}
\subsubsection{Depth Prediction Transformer (DPT)}
The DPT model utilizes a Vision Transformer (ViT) as its backbone, which processes images in a sequence of patches. Unlike standard transformers, DPT assembles tokens from different stages of ViT into multi-resolution feature maps, which are then progressively fused using a convolutional decoder to produce full-resolution predictions. This method ensures that the model captures both fine details and global context effectively. The model operates on a high constant resolution throughout the process, enhancing its ability to generate detailed and coherent predictions across the entire image.

In this study, the DPT was directly applied to high-resolution RGB images collected from commercial apple orchards to generate detailed RGB-D (depth) data. The depth maps produced by DPT provided the 3D spatial context necessary for validating the pose estimation performed by the YOLO11 algorithm. Specifically, major axes of immature green fruits estimated by YOLO11 pose estimation model were compared against measurements obtained from 3D point clouds generated by DPT. The actual diameters of the fruitlets were also measured in the field using digital calipers to provide ground truth validation.

High-resolution RGB images were captured using specialized cameras mounted on a robotic platform as discussed before. These images were then preprocessed to enhance their quality, which were then used as inputs into the DPT model. The model processed these images through several stages, including patch extraction, transformer encoding, thresholding and feature map creation, resulting in RGB-D outputs. Subsequently, the model output was used to determine 3D coordinates of the key pints (opposite ends of the longitudinal axis of fruitlets) estimated by YOLO11 pose estimation model, which resulted in 3D poses and major axis lengths of the fruitlets. The estimated major axis length was then compared against the field measurements as discussed below.

\subsubsection{Depth Anything V2}
Depth Anything V2 is an advanced monocular depth estimation model designed to convert standard RGB images into depth-mapped (RGB-D) images \cite{yang2024depth} . This model employs a fully convolutional network to interpret and transform visual data (on single RGB images) into spatial mappings.  The architecture processes images through multiple layers to extract and refine features at various scales, capturing detailed and granular elements ideal for complex natural scenes. Utilizing an encoder-decoder network, the model compresses the image into a feature-rich format, and then expands these features to assign depth values across the image. Enhanced by skip connections that preserve spatial details, the model produces a dense depth map that accurately reflects the real-world geometry of the scene.

Similar to DPT, the Depth Anything V2 model was also utilized to estimate 3D locations of two key points of fruitlets (opposite ends of fruitlet's major axis) estimated by the YOLO11 model as discussed before which is crucial before the thinning process in agricultural operations. As depicted in Figure \ref{fig:bothVits}b, the original RGB images (same images with 70 sample fruitlets mentioned before) collected using a consumer-grade camera were provided as inputs to Depth Anything V2 model. The images then went through processes including feature extraction, feature compression through an encoder to create a dense representation, and expanded representation with a decoder to generate detailed depth maps. The integration of skip connections helped preserve essential spatial details. The resulting depth maps provided comprehensive 3D spatial data, which was used to estimate the 3D coordinates of key points estimated by the YOLO11 model.

\subsection{Field validation of estimated pose}
\begin{figure}[ht]
\centering
\includegraphics[width=0.8 \linewidth]{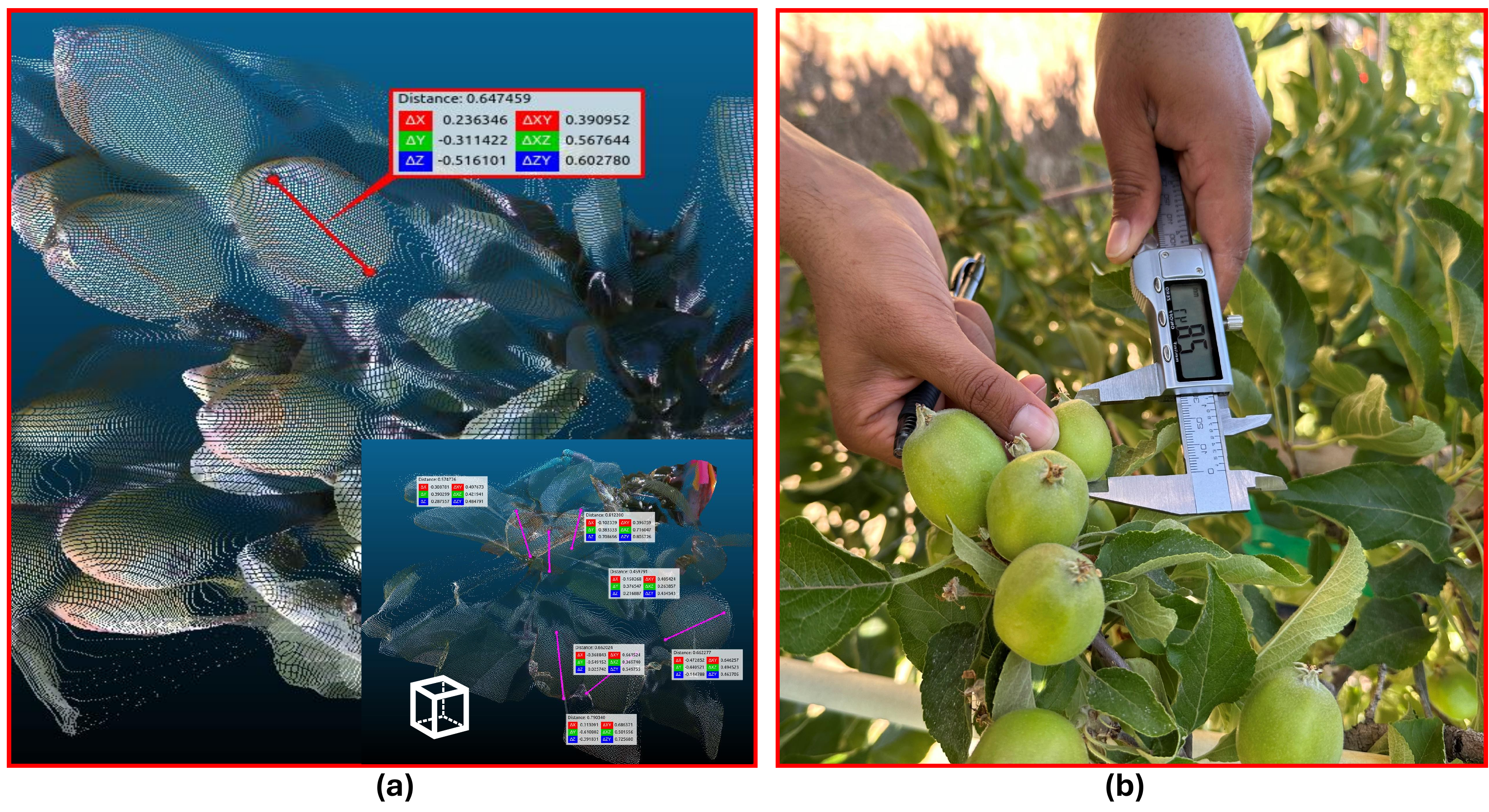}
\caption{Field-Level Validation of detected Pose : (a) displays the 3D point clouds derived from RGB to RGB-D mapping using Vision Transformers ;  (b) shows the vernier caliper used for ground truth measurements.}
\label{fig:Cloudcompare}
\end{figure}

RGB-D maps generated by two Vision Transformer models were used to extract resulting 3D point clouds (Figure \ref{fig:Cloudcompare})a. As discussed before, the image-based key points of fruitlets (two opposite ends of the fruitlet's major axis) were converted into 3D coordinates in physical units using the point clouds generated by the transformer models as well as the point cloud captured by the RGB-D camera for comparison. Once 3D coordinates of the endpoints of the fruitlets were estimated, the distance between them was calculated (using a Euclidean distance formula) as the fruitlet size. To assess the accuracy of the estimated fruitlet sizes based on three different 3D point cloud datasets (two Vision Transformer generated and one sensor generated), ground truth fruit size (major axis length) was measured in the filed using a Vernier Caliper as illustrated in Figure \ref{fig:Cloudcompare}b. Altogether, field measurements were collected for 70 randomly selected samples of immature green fruitlets. The accuracy assessment of fruitlet size estimation conducted with the in-field measurements served as a benchmark for validating the fruitlet poses estimated by the integrated YOLO11 and Vision Transformer models that were employed to estimate fruitlet key points and transform standard RGB images into detailed RGB-D data without additional depth information from the camera.

The evaluation of the predicted fruit size or major axis length against actual field measurements was performed using CloudCompare (Seattle, Washington, USA), a tool for assessing the quality of 3D point clouds. Root Mean Square Error (RMSE) and Mean Absolute Error (MAE) were the primary metrics used to evaluate the accuracy of the pose estimations, specifically measuring discrepancies in the length of green fruits. The RMSE is defined as:

\begin{equation}
\text{RMSE} = \sqrt{\frac{1}{n} \sum_{i=1}^{n} (\text{Predicted Length}_i - \text{Actual Length}_i)^2}
\label{eq:rmse}
\end{equation}

and the MAE as:

\begin{equation}
\text{MAE} = \frac{1}{n} \sum_{i=1}^{n} \left| \text{Predicted Length}_i - \text{Actual Length}_i \right|
\label{eq:mae}
\end{equation}

where \textit{n} is the number of fruits analyzed, \textit{Predicted Length} is the major axis length or size of the fruitlet estimated using the depth maps, and \textit{Actual Length} is the major axis length or size of the fruitlet measured in the field using a caliper. These metrics, RMSE and MAE, provided a quantitative measure of the model's precision in estimating the pose of immature green fruits based on their size, facilitating an objective comparison between the predicted and actual dimensions.

\section{Results and Discussion}
\subsection{Fruitlet Detection and Pose Estimation}
\par The comparative assessment of YOLO11 and YOLOv8 models, as presented in Table \ref{tab:metrics1}, showed similar performance of YOLO11 and YOLOv8 configurations in terms of precision, with YOLO11n achieving the highest precision of 0.91 and YOLOv8m achieving 0.90. In terms of recall as well, these two configurations performed similarly with YOLOv8n achieving 0.905 and YOLO11x achieving 0.891, demonstrating their strength in capturing all relevant instances. However, the Table \ref{tab:metrics1}, showed varying performance of different models in terms of specific performance measures, which emphasizes the importance of selecting the appropriate model configuration based on the specific needs of real-time pose estimation tasks.
\begin{table*}[ht]
\centering
\caption{Precision and recall metrics for box and pose estimations of apple fruitlet achieved with for YOLO11 and YOLOv8 configurations. The table focuses on comparing the models with these metrics excluding mAP@50, which is discussed separately.}
\label{tab:metrics1}
\begin{tabular}{|c|c|c|c|c|}
\hline
{\textbf{Model Configurations}} & \multicolumn{2}{c|}{\textbf{Box Detection}} & \multicolumn{2}{c|}{\textbf{Pose Estimation}} \\ \cline{2-5} 
                                & \textbf{Precision} & \textbf{Recall} & \textbf{Precision} & \textbf{Recall} \\ \hline
YOLO11n                         & \textbf{0.91}               & 0.872           & \textbf{0.915}              & 0.889           \\ \hline
YOLO11s                         & 0.891              & 0.877           & 0.899              & 0.891           \\ \hline
YOLO11m                         & 0.887              & 0.864           & 0.91               & 0.887           \\ \hline
YOLO11l                         & 0.866              & 0.877           & 0.884              & 0.894           \\ \hline
YOLO11x                         & 0.889              & 0.891           & 0.903              & 0.905           \\ \hline
YOLOv8n                         & 0.86               & \textbf{0.905}           & 0.877              & 0.925           \\ \hline
YOLOv8s                         & 0.883              & 0.875           & 0.896              & 0.895           \\ \hline
YOLOv8m                         & 0.908              & 0.838           & 0.879              & 0.887           \\ \hline
YOLOv8l                         & 0.843              & 0.69            & 0.848              & 0.706           \\ \hline
YOLOv8x                         & 0.896              & 0.886           & 0.888              & \textbf{0.91}           \\ \hline
\end{tabular}
\end{table*}

\begin{figure*}[ht]!
\centering
\includegraphics[width= 0.7\linewidth]{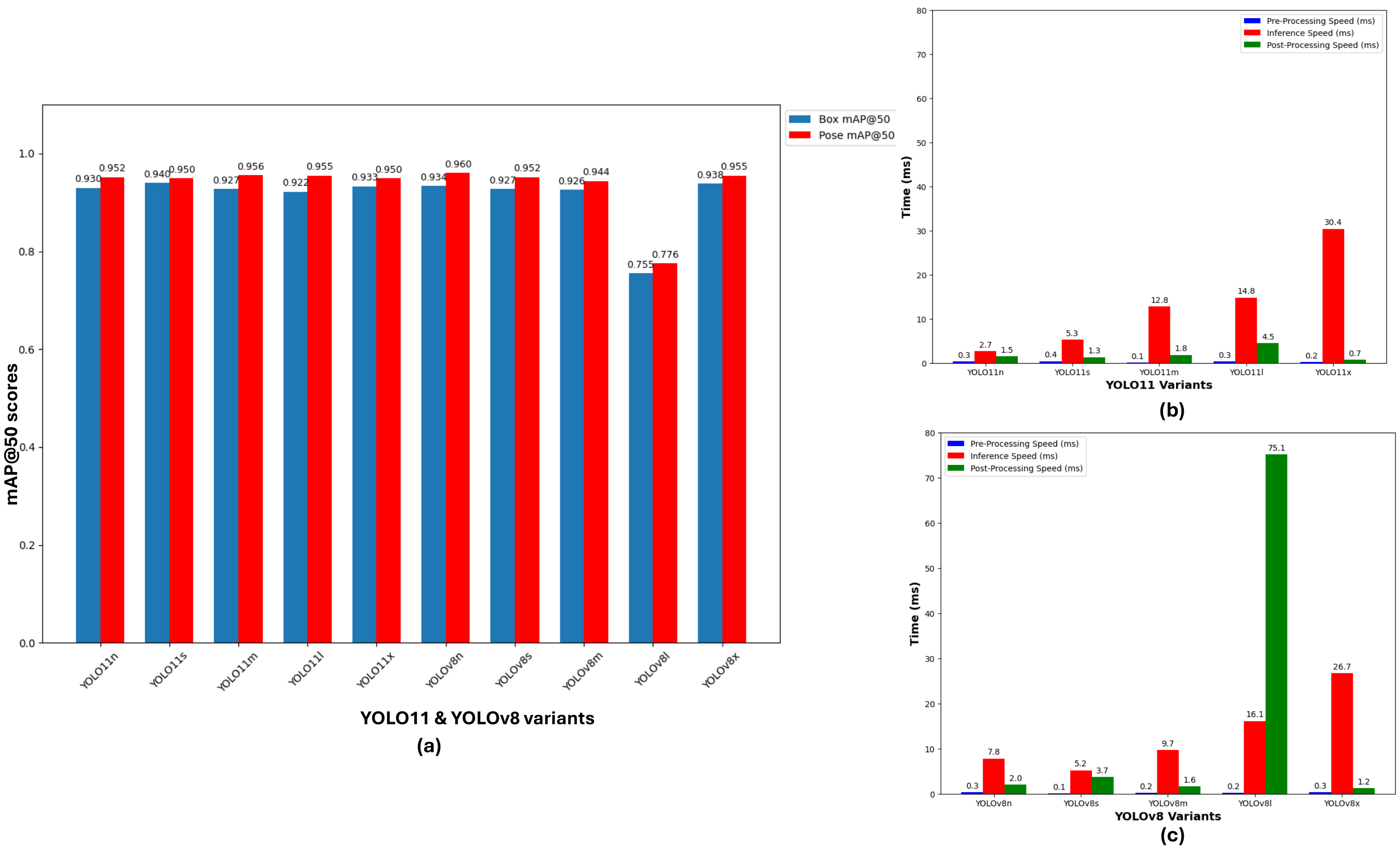}
\caption{Additional performance metrics for YOLO11 and YOLOv8 configurations; (a) mean Average Precision at 50\% mAP@50; (b)  Image processing time (pre-processing, inference, and post-processing) for YOLO11 configurations; and (c) Image processing time (pre-processing, inference, and post-processing) for YOLOv8 configurations}
\label{fig:MAPandInference}
\end{figure*}

Figures \ref{fig:MAPandInference}a illustrated mean Average Precision at 50\% Intersection over Union (mAP@50) values of all the model configurations tested. As in the case of Precision and Recall, the YOLO11 and YOLOv8 models achieved similar mAP@50), with YOLO11s achieving 0.94 and YOLOv8n slightly behind at 0.934, demonstrating the comparable effectiveness of both models in complex agricultural settings where precision, recall, and mAP@50 are crucial. However, there was a substantial difference in the computational efficiency of YOLO11 and YOLOv8 model configurations, thus appearing as a decisive factor in model selection. Figures \ref{fig:MAPandInference}b and \ref{fig:MAPandInference}c present the image processing speeds, showcasing YOLO11's faster processing speed compared to the same for YOLOv8 model configurations. Particularly, YOLO11n was the best model configuration in terms of inference speed with 2.7 milliseconds processing time per image, which was significantly better than the quickest YOLOv8 configuration, YOLOv8n, which took 7.8 milliseconds. This considerable difference in inference speeds is pivotal, especially in scenarios demanding rapid data processing such as real-time agricultural operations. The faster processing capabilities of YOLO11n, coupled with its competitive precision and recall, makes it the preferred model for our robotic thinning project. These results not only highlight the model's robustness in detecting and estimating the pose of immature green fruits but also affirm its potential to significantly enhance operational efficiency and accuracy in the field.

\par
\begin{figure*}[ht!]
\centering
\includegraphics[width= 0.95\linewidth]{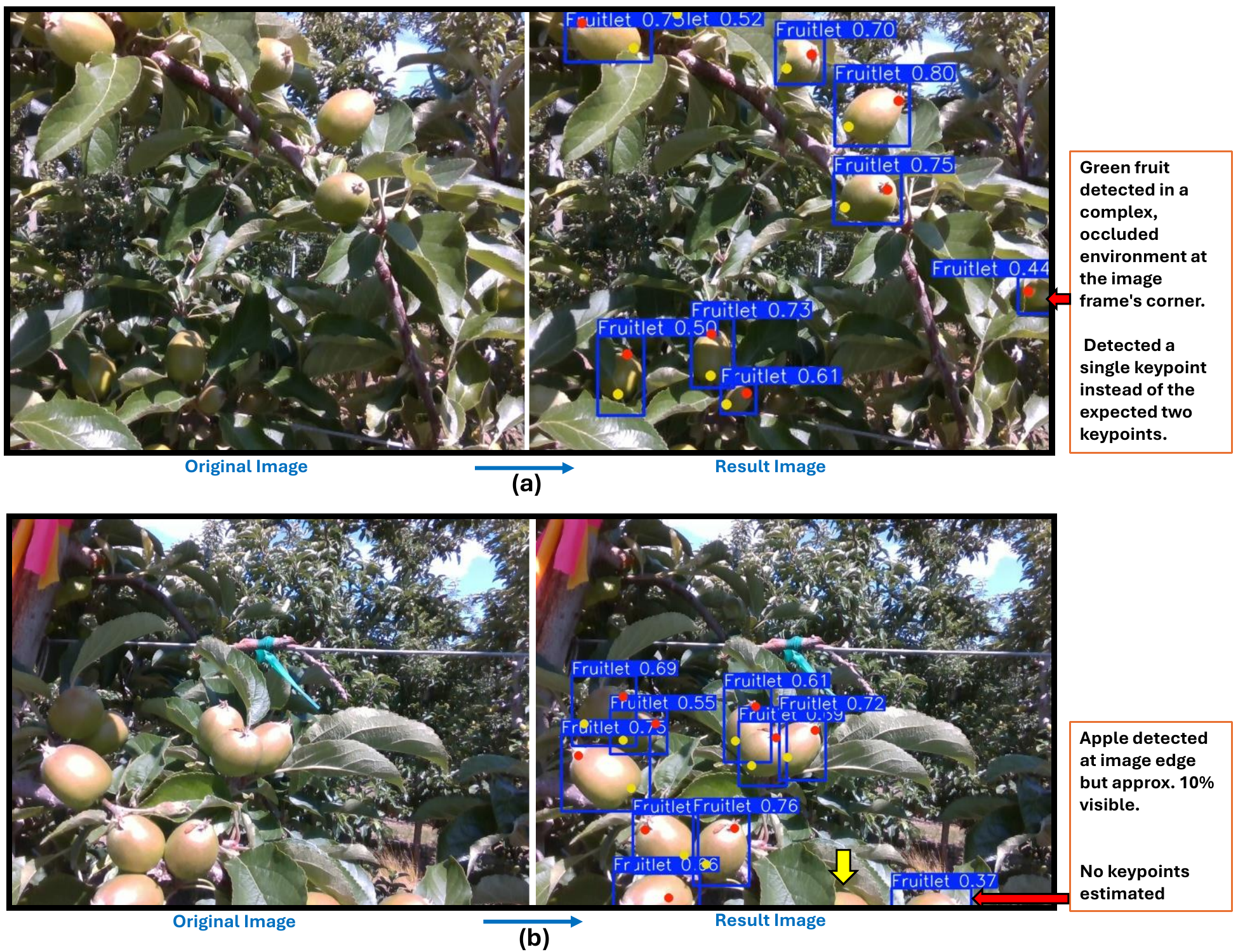}
\caption{Demonstrationg the YOLO11n model's effectiveness in detecting and estimating the pose of immature green fruits within a commercial orchard setting: (a) Showing accurate fruit detection in shadowed and low-light conditions, emphasizing the model's adaptability ; (b) Showing the detection of immature green fruits with minimal visibility, demonstrating the model's sensitivity to subtle visual cues and its robust performance in complex agricultural environments.}
\label{fig:ResultImage1}
\end{figure*}

 In the results depicted in Figure \ref{fig:ResultImage1}, the examples of the YOLO11n model in the detection and pose estimation of immature green fruits within a commercial orchard environment is illustrated. The immature green fruits, even those occluded by shadows or canopy foliage, were robustly identified, as evidenced in Figure \ref{fig:ResultImage1}a. In this subfigure, a green fruit located in a shadowed and low-light area was accurately detected, an ability highlighted by the red arrow, indicating the model's robust performance under diverse environmental conditions.

Furthermore, within the bounding boxes of detected fruits, keypoints were identified by the model, although not all keypoints could be detected when parts of the fruit were not visible, demonstrating a realistic limitation given the partial visibility of some fruits. Such instances are portrayed in Figures \ref{fig:ResultImage1}a and \ref{fig:ResultImage1}b, showcasing the YOLO11n model's consistent accuracy in detection and pose estimation. Remarkably, the model also identified fruits that were only marginally visible; an example is depicted in Figure \ref{fig:ResultImage1}b, where approximately 10-15\% of an immature green fruit at the edge of the image frame was detected. This capability emphasizes the YOLO11 model's refined sensitivity to minimal visual cues, asserting its robustness in complex agricultural settings. 

\begin{figure*}[ht]!
\centering
\includegraphics[width= 0.7\linewidth]{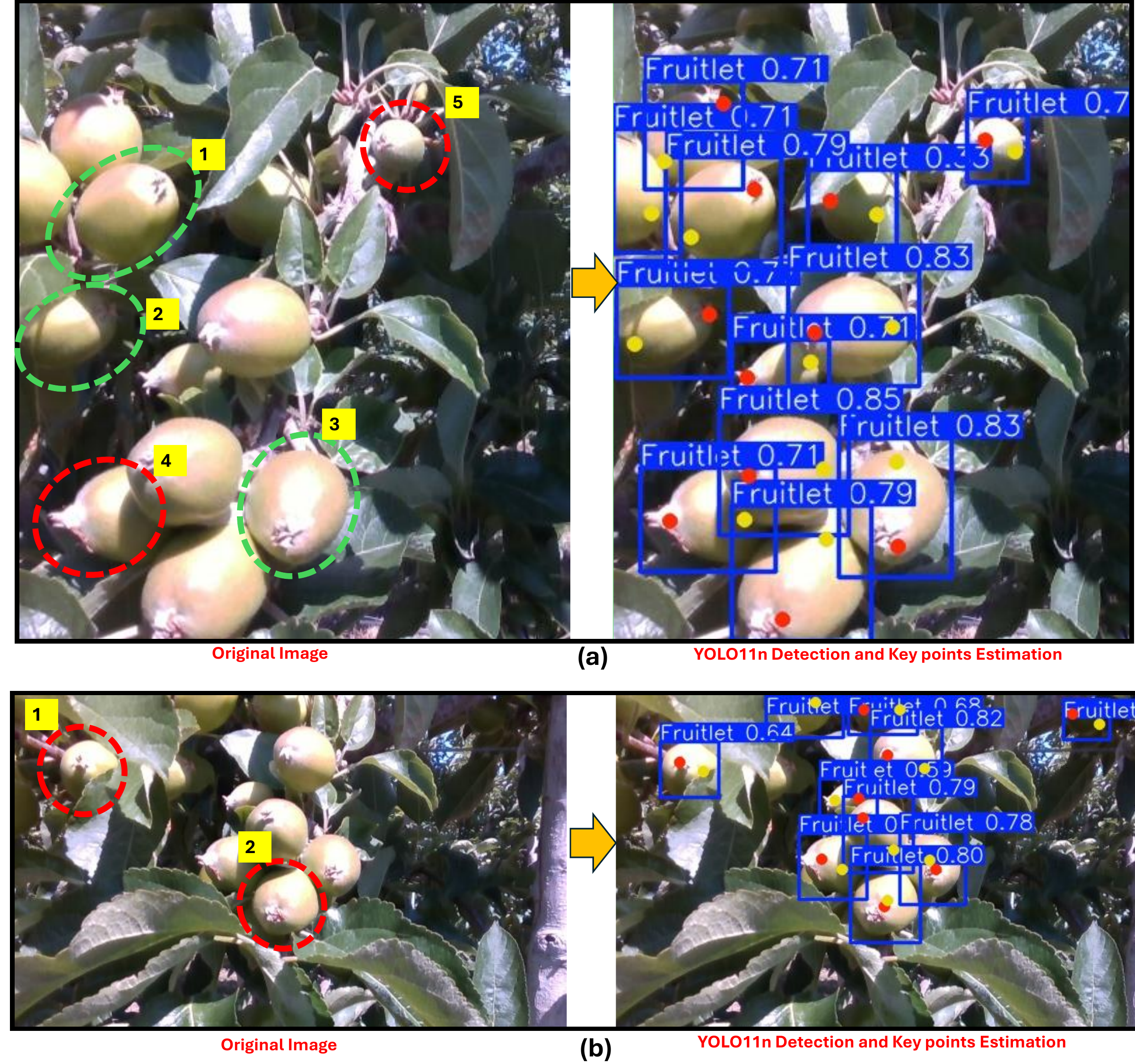}
\caption{Demonstration of discussion on the YOLO11n model's detection and pose estimation capabilities in a commercial orchard: (a) Showing the successful detections within densely clustered environments, highlighting the model's robust performance despite occlusions; (b) Showing the challenges in pose estimation due to fruit orientation, pointing to areas for future algorithmic improvements.}
\label{fig:ResultsYOLO11nDetectionPose}
\end{figure*}

Additionally, in Figure \ref{fig:ResultsYOLO11nDetectionPose}, further examples of the YOLO11n model's detection and pose estimation capabilities for immature green fruits are demonstrated. The left side of each image in the figure displays the original scenes, while the right side presents the detection and pose estimation results by YOLO11n. Although the model performed commendably in terms of detection and pose estimation, areas for improvement were identified.

For instance, in Figure \ref{fig:ResultsYOLO11nDetectionPose}a, the depiction of the detection and pose estimation in a densely clustered environment of immature green fruits reveals critical challenges. In the heavily clustered setting, which is typical before thinning operations in commercial 'Scifresh' orchards, the model managed to detect the majority of immature green fruits. Notably, strong pose detection was observed, particularly in the green-circled regions of the left side, which corresponds to the original imagery.

However, the red-circled regions on the right side of Figure \ref{fig:ResultsYOLO11nDetectionPose}a present a limitation. While the model detected fruits and estimated keypoints up to the visible sections of the fruits' major axes, the accuracy of these estimations might be compromised by occlusions. Specifically, the fruit positioned above the red-circled apple complicates the detection scenario.

For robotic thinning operations, a technical solution would involve prioritizing the thinning of fruits that occlude others to enhance visibility and accuracy in subsequent detections. This approach would improve the practicality of robotic systems designed for fruitlet thinning, enabling more effective deployment in commercial orchard environments. Future research should focus on developing algorithms that can adaptively prioritize detection and thinning sequences based on occlusion and fruit arrangement, enhancing the efficacy of robotic thinning operations. 

Likewise, in the upper part of Figure \ref{fig:ResultsYOLO11nDetectionPose}a, particularly within the red circle, the orientation of an immature green fruit is captured in an orientation that predominantly displays its minor axis rather than the major axis from calyx to peduncle. The YOLO11 model successfully identified the presence of the fruit, a demonstration of its robust detection capabilities. However, the pose estimation within the bounding box, as depicted on the right side of the figure, did not accurately reflect the major axis of the fruit. This inaccuracy is understandable, given the natural complexity and density of fruitlets within apple orchards. Apple trees often produce a surplus of fruitlets and flowers, far exceeding what is necessary for optimal fruit development, as highlighted in historical agricultural studies \cite{dennis2000history, costa2018fruit}. Such conditions pose significant challenges for pose estimation algorithms, particularly in scenarios where the fruit's orientation does not align with the typical major axis configurations. 

Moreover, in the depicted results of \ref{fig:ResultsYOLO11nDetectionPose}b, the YOLO11n model accurately recognized the object as an immature green apple; however, the pose estimation within the bounding box revealed inaccuracies. Although one keypoint (at the calyx) was correctly identified, the other necessary for a complete pose estimation (at the peduncle) was obscured due to occlusion, a persistent challenge in dense orchard environments. This partial visibility compromises the effectiveness of even the most advanced models like YOLOv10, which demonstrated the highest precision in our study. The upper part of \ref{fig:ResultsYOLO11nDetectionPose}b illustrates this limitation: the visible calyx, while detected, does not provide sufficient information to infer the full fruit pose. Future work might benefit from integrating machine learning technologies such as Generative Adversarial Networks (GANs), which have shown potential in reconstructing complete object profiles from limited views \cite{lu2022generative}. Such approaches could potentially address the occlusions inherent in agricultural settings by extrapolating the unseen portions of fruits, enhancing the accuracy and utility of robotic thinning tools in commercial orchards.

Furthermore, the lower portion of the red-circled area in Figure \ref{fig:ResultsYOLO11nDetectionPose}b presents another challenge where the YOLO11n model, despite its accurate detection of immature green fruits, failed to correctly identify the keypoints for pose estimation. This limitation was notably influenced by the insufficient diversity and volume of the training dataset used in this study, which comprised only 1147 images. The lack of representative data from densely occluded or complex environments may hinder the model's learning algorithm, leading to less accurate pose estimation in real-world scenarios. Enhancing the training dataset with a broader array of images capturing a wider range of fruit orientations and occlusion levels could significantly improve the model's ability to generalize and perform more reliably in commercial orchard settings. Such an expansion of the dataset is anticipated to provide the necessary variability that would enable the model to better learn and predict fruit poses accurately, thereby supporting more effective robotic applications in agriculture.

\subsection{Evaluation of Vision Transformers in Pose Length Validation (Fruit's Major Axis Length) } The assessment of major axis length, an indicator of pose accuracy, was rigorously conducted to evaluate the performance of Vision Transformers in the context of enhanced fruit size estimation. This analysis combined the detection capabilities of YOLO11 with the depth estimation precision of Vision Transformers. The validation process utilized data from Intel RealSense point clouds, the Dense Prediction Transformer (DPT), and the Depth Anything model, all benchmarked against ground truth measurements of immature green fruit lengths. Initially, data from Intel RealSense displayed substantial deviations from actual measurements, with a Root Mean Square Error (RMSE) of 9.9845 and a Mean Absolute Error (MAE) of 7.7444. The application of the DPT model markedly improved accuracy, reducing the RMSE to 3.2935 and the MAE to 2.6237. However, the most significant enhancements in estimating the major axis length were observed with the Depth Anything model, which demonstrated the lowest errors, achieving a RMSE of 1.5261 and an MAE of 1.2809. These results highlight the effectiveness of integrating YOLO11 with advanced visual transformers like DPT and Depth Anything in refining the precision of pose length estimations for agricultural applications, particularly in robotic fruit thinning operations.
\begin{figure*}[ht!]
\centering
\includegraphics[width= 0.75 \linewidth]{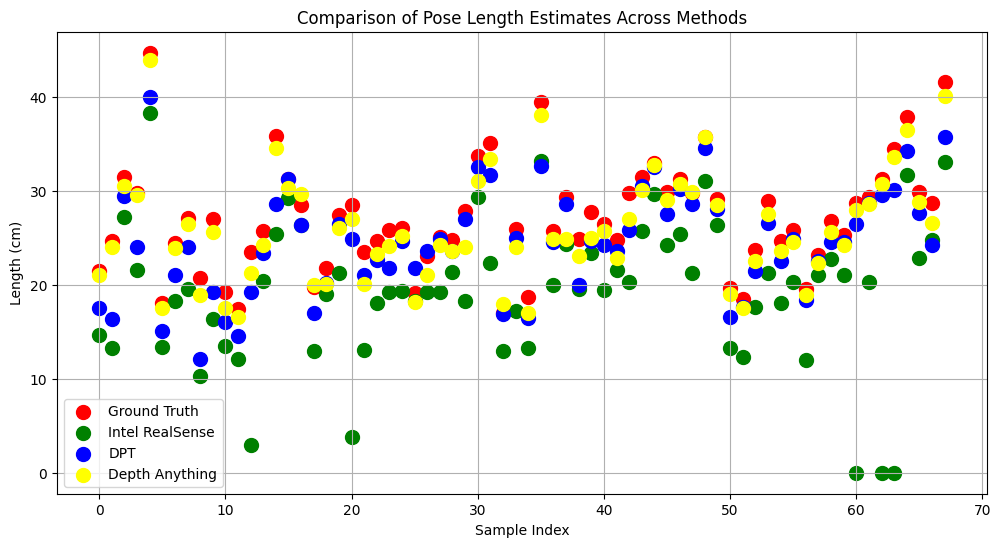}
\caption{Box plot comparing pose length estimates for 70 samples  where 'Ground Truth' represents direct field measurements, 'Intel RealSense' shows lengths derived from camera-based point clouds, 'DPT' (Dense Prediction Transformer) and 'Depth Anything' illustrate lengths from visual transformer-generated depth maps. }
\label{fig:PoseValidation1}
\end{figure*}

\begin{figure*}[ht!]
\centering
\includegraphics[width= 0.75 \linewidth]{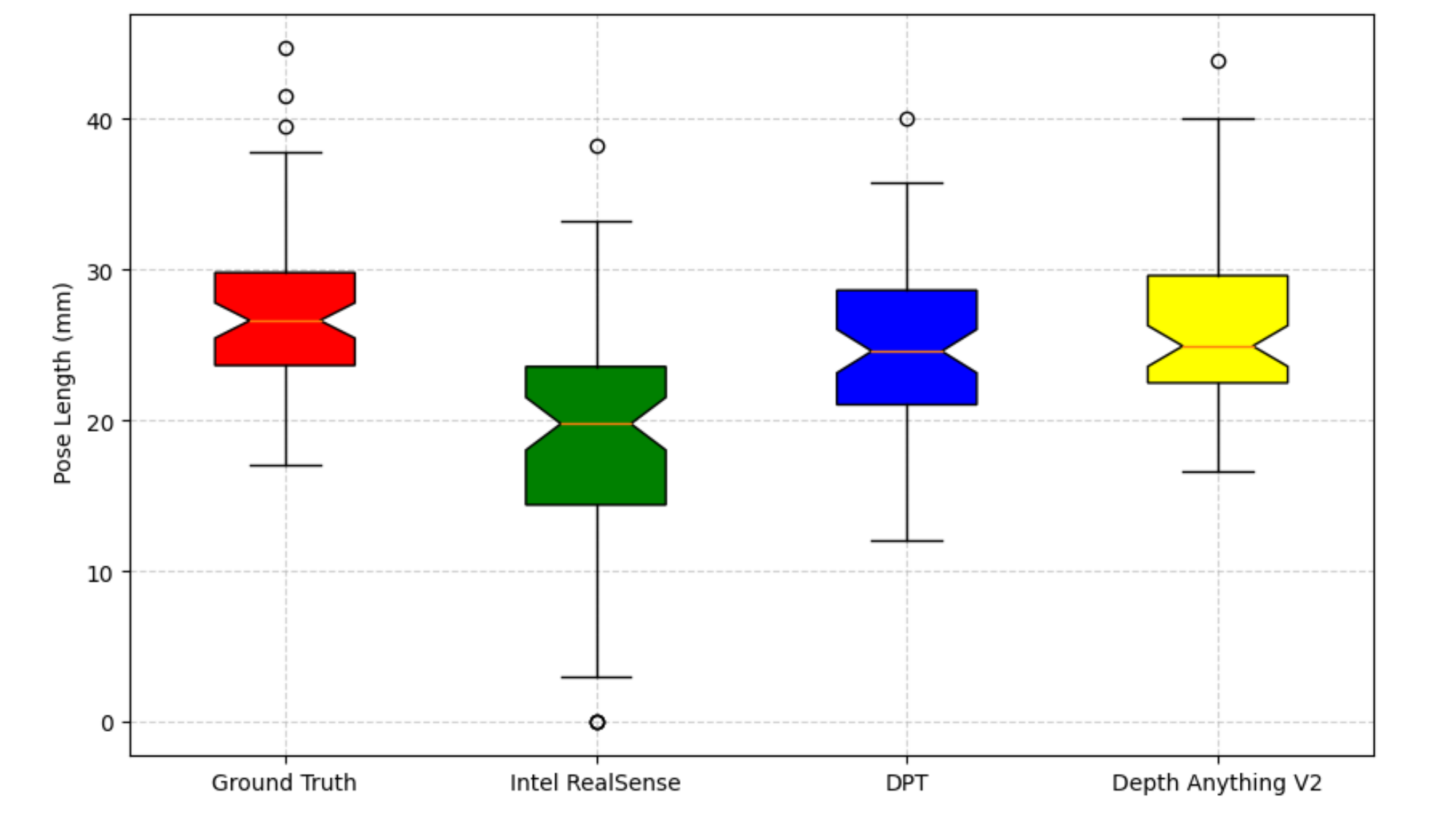}
\caption{Box plot comparing pose length estimates for 70 samples  where 'Ground Truth' represents direct field measurements, 'Intel RealSense' shows lengths derived from camera-based point clouds, 'DPT' (Dense Prediction Transformer) and 'Depth Anything' illustrate lengths from visual transformer-generated depth maps. }
\label{fig:PoseValidation}
\end{figure*}

\begin{figure*}[ht!]
\centering
\includegraphics[width= 0.75 \linewidth]{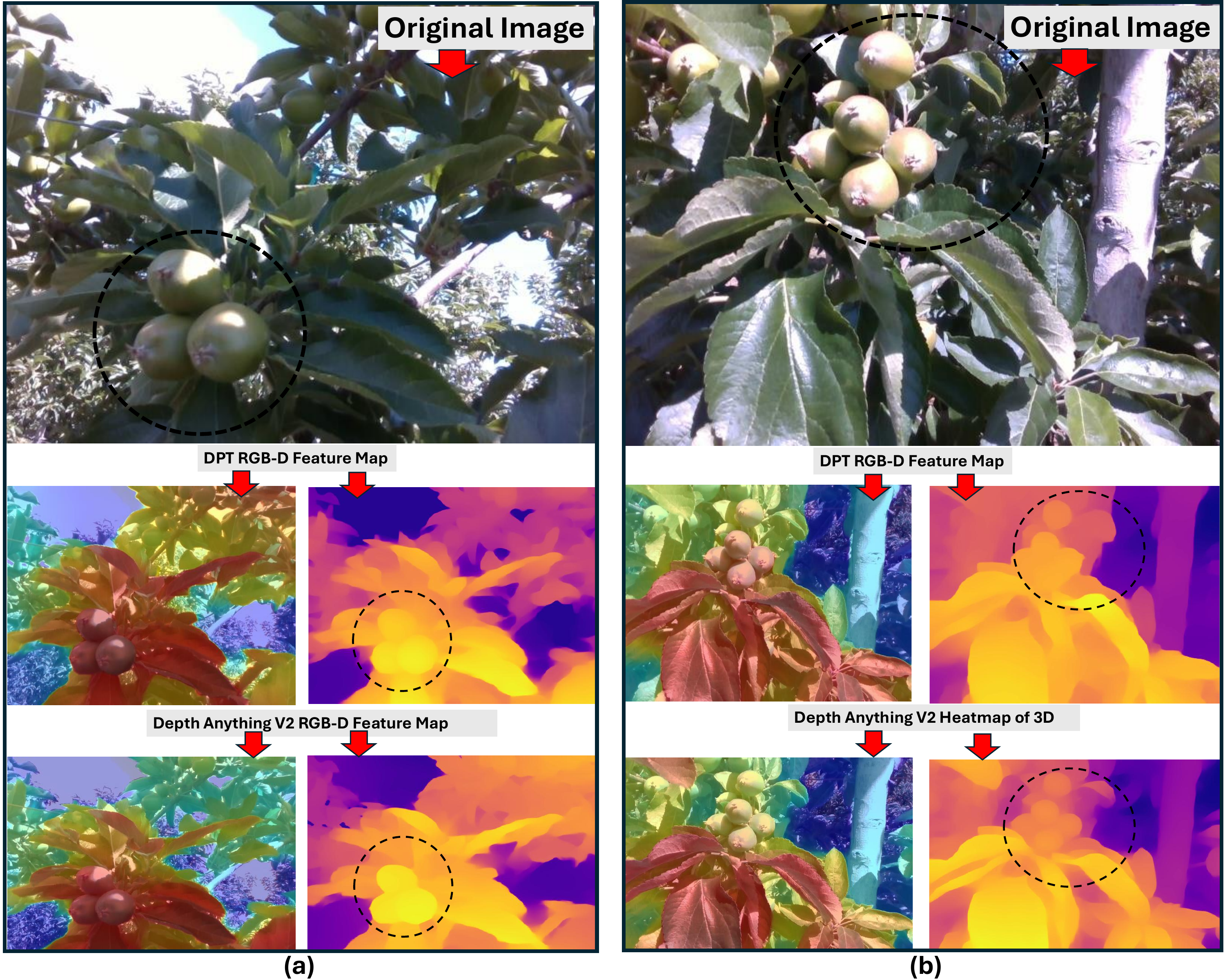}
\caption{Illustrating the analysis of densely clustered immature green fruits using visual transformers: (a) DPT and (b) Depth Anything. Top layers display original RGB images, followed by RGB-D feature maps, depth heatmaps, and extracted point clouds, highlighting the depth processing and spatial analysis capabilities of each transformer.}
\label{fig:ResultsVisionTransformers}
\end{figure*}

The utilization of visual transformers for RGB to RGB-D mapping is exemplified in the Figure \ref{fig:ResultsVisionTransformers}, which shows examples of DPT in Figure \ref{fig:ResultsVisionTransformers}a and Depth Anything V2 in Figure \ref{fig:ResultsVisionTransformers}b. 
In Figure \ref{fig:ResultsVisionTransformers}a, a notable observation in the black circled region of the original RGB image reveals three fruits, one of which is partially illuminated by sunlight, impacting the depth perception in the generated images. In the corresponding feature RGB-D maps for both DPT and Depth Anything, although the structural details are well-preserved, it is the Depth Anything model that demonstrates superior refinement. The 3D point cloud heatmap from Depth Anything distinctly captures the intricate structure of the canopy foliage and the spatial arrangement of the fruits with greater clarity and detail compared to DPT. This is particularly evident in the enhanced delineation of the fruit affected by the sunlight, where Depth Anything accurately models the depth variations caused by the lighting conditions. This figure serves as a compelling demonstration of Depth Anything's advanced capabilities over DPT in processing complex visual data from agricultural environments. The refined depth mapping provided by Depth Anything is instrumental for applications in green fruit automation, where precise spatial data is crucial for effective robotic thinning.

Likewise, Figure \ref{fig:ResultsVisionTransformers}b, focuses on a densely packed cluster of seven immature green fruits, a typical scenario crucial for robotic thinning. The original RGB image, highlighted in black dotted circles, showcases the high density of fruitlets, emphasizing the necessity for precise thinning operations. Depth Anything V2's performance in this intricate setting is particularly notable. The model's output, as reflected in the heatmap of the 3D point clouds, reveals an exceptional level of detail not only in identifying the target fruits but also in mapping the surrounding canopy foliage (Figure \ref{fig:ResultsVisionTransformers}b). This enhanced depth perception by Depth Anything is crucial for robotic applications, where distinguishing between closely clustered fruits and their environmental context is vital for effective operation. The superior depth mapping capabilities of Depth Anything, compared to DPT, are distinctly visible in the black dotted regions of Figure b. It successfully delineates the spatial relationships and depth variations within the cluster, providing a clearer understanding of the physical layout essential for robotic maneuvering and decision-making in thinning operations. This example underscores Depth Anything V2's robustness and accuracy in handling complex agricultural environments, making it a valuable tool for advancing the precision and efficiency for robotic fruit thinning technologies.

\subsection{Discussion}
\begin{figure*}[ht!]
\centering
\includegraphics[width= 0.85 \linewidth]{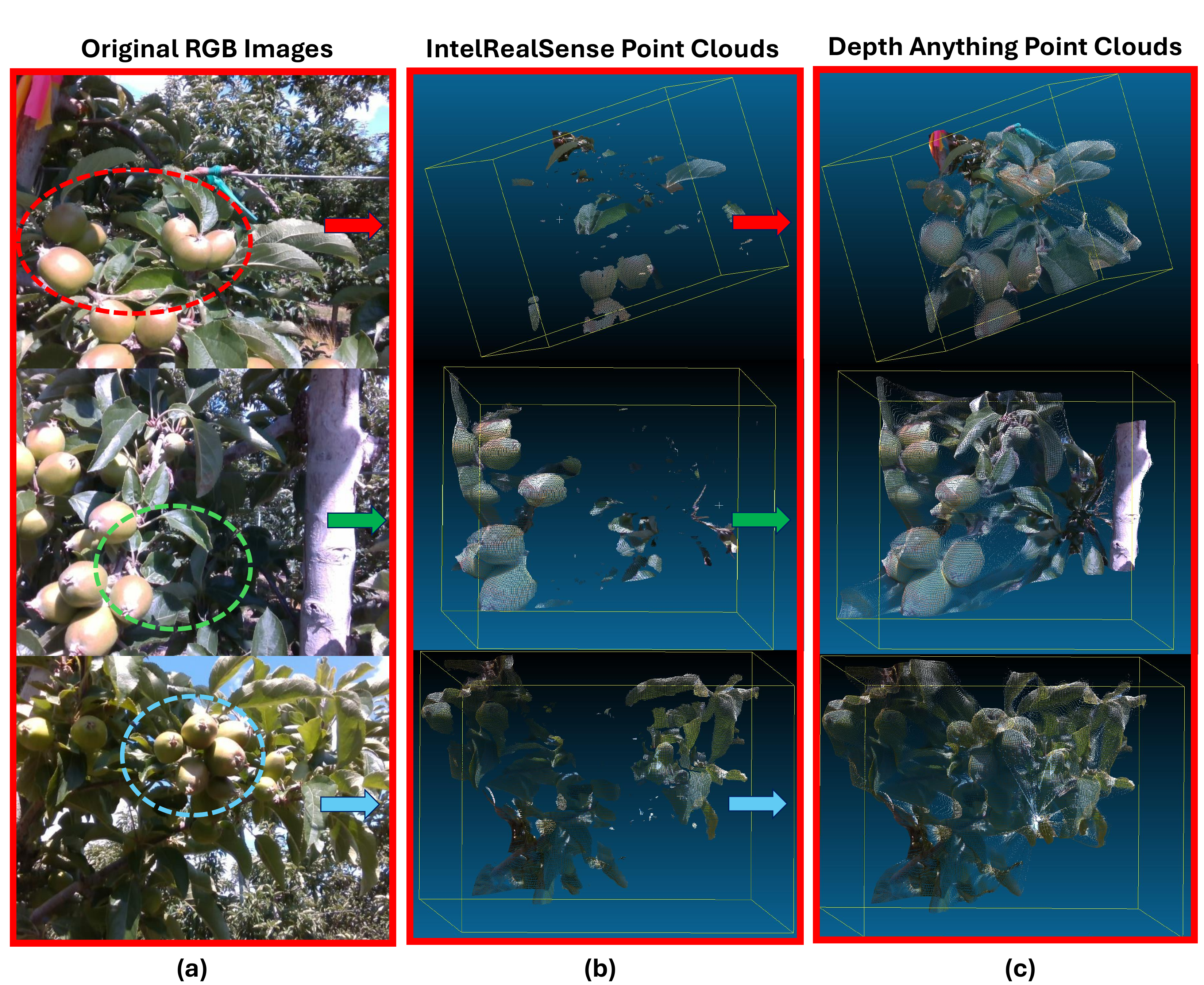}
\caption{Comprehensive visualization of depth estimation techniques for immature green fruit pose analysis: (a) Original RGB images depict the initial visual state of densely populated orchard scenes; (b) RealSense Point Clouds illustrate depth information captured via RealSense technology; (c) Depth Anything Model Point Clouds showcase enhanced depth accuracy, outperforming DPT in RMSE and MAE metrics, critical for precise pose estimation.}
\label{fig:RGBDresults}
\end{figure*}

The integration of 3D data is essential in robotics for efficient and effecive navigation and manipulation of target objects \cite{hagele2016industrial, amparore2022robotic}. Numerous studies have emphasized that without robust 3D information, effective automation remains unattainable \cite{barazzetti2010orientation,fathi2015automated,dadhich2016key, tian2020computer,  xia2015situ, vougioukas2019agricultural, he2018sensing}, highlighting the critical role of precise spatial data in enhancing robotic functionalities. The utilization of the Depth Anything V2 transformer, in particular, has proven highly effective in generating detailed 3D mappings from standard RGB images, as illustrated in Figure \ref{fig:RGBDresults}. Subfigure \ref{fig:RGBDresults}a presents the original RGB images captured in a Scifresh orchard, while Subfigure \ref{fig:RGBDresults}b displays the corresponding Intel RealSense point clouds, and Subfigure \ref{fig:RGBDresults}c showcases the Depth Anything V2 generated point clouds. These point clouds were produced at a data collection threshold of 2 feet, demonstrating the transformer's capacity to accurately interpret and convert 2D images into spatially informative 3D data. This transformation process is crucial for tasks that require precise depth perception, such as robotic fruitlet thinning in densely populated orchards. By providing a clear 3D representation of the environment, Depth Anything V2 facilitates more accurate and efficient robotic interactions, significantly reducing the operational complexity and enhancing the efficacy of automated procedures. The results presented highlight the potential of these advanced vision transformers to not only improve the accuracy of automated operations in agricultural settings but also to drive innovations that could revolutionize the sector.

In the exploration of RGB to RGB-D mapping through vision transformers, particular attention is drawn to Figure \ref{fig:RGBDresults}a where specific regions marked by red, green, and blue dotted circles are critical. These areas, highlighted across the top, middle, and bottom images, contain key fruitlets and essential canopy features integral to the operational success of future robotic applications like fruitlet thinning and canopy management. The associated point clouds, as depicted in Figure \ref{fig:RGBDresults}b, extracted directly from the Intel RealSense machine vision camera, unfortunately, do not adequately capture the detailed structure of fruitlets and canopy elements within these circled regions. Interestingly, these missing details in the camera-captured point clouds are effectively compensated for by the Depth Anything vision transformer model, as evidenced in Figure \ref{fig:RGBDresults}c. This model adeptly generates the necessary 3D data for areas where the camera-based point clouds fall short, ensuring that even obscured or partially visible fruitlets are accounted for. This capability of Depth Anything not only enhances the precision of 3D mapping but also supports more accurate robotic interactions with the environment by providing a more comprehensive spatial understanding.

In the middle section of subfigures in Figure  \ref{fig:RGBDresults}, a critical observation regarding the tree trunk—a vital element for collision avoidance and robotic navigation in orchard settings is highlighted. Due to excessive sunlight exposure or possible intrinsic camera limitations, the Intel RealSense camera failed to capture depth information for the tree trunk on the rightmost part of the image, as shown in Figure  \ref{fig:RGBDresults}b. Remarkably, this missing data was effectively reconstructed by the Depth Anything vision transformer, which successfully generated the requisite point clouds for this region. This capability underscores the robustness and utility of Depth Anything in enhancing environmental perception, crucial for the safe and efficient operation of agricultural robots.

In past several studies that have explored pose estimation in various agricultural contexts, a broad spectrum of approaches has been highlighted along with their respective challenges. For instance, Eizentals et al. \cite{eizentals20163d} employed a Lidar-based method for green peppers which, while effective under controlled settings, underscores the limitations inherent in relying on precise laser range findings for cluttered outdoor environments. Similarly, methods leveraging RGB-D sensors \cite{lin2019guava} and binocular imagery \cite{yin2021fruit} have demonstrated high precision in fruit detection and pose estimation. However, these approaches often suffer from prolonged processing times and complex setups, hindering their feasibility for real-time applications. Further, studies like those by Sun et al. \cite{sun2023citrus} and Li et al. \cite{li2018pose} have introduced innovative techniques using single RGB images and symmetry detection in point clouds, respectively. While these methods show promise, they are constrained by their specific requirements for image annotations and environmental conditions, which may not be consistently replicable in dynamic field settings. Moreover, recent advancements by Kim et al. \cite{kim20232d} and Giefer et al. \cite{giefer2019deep} in 2D pose estimation and apple rotation regression highlight the ongoing need for methods that can robustly handle occlusions and offer versatility across different orchard environments. In contrast to these computationally intensive methods, our study harnesses straightforward, low-cost RGB imaging techniques to reconstruct RGB-D data with high accuracy, significantly simplifying the data collection and processing pipeline. By implementing a critical threshold for data input into vision transformers like Depth Anything, we achieve refined 3D mappings essential for precise robotic operations. This approach not only reduces the computational load but also enhances the scalability and applicability of pose estimation technology in real-world agricultural settings, offering a more practical solution for field-level validation and operational efficiency.

\section{Conclusion}
This research has highlighted the growing interest and advancements in robotic solutions for labor-intensive agricultural tasks, such as the thinning of immature green fruits in commercial apple orchards. While consumer-level automation for harvesting red apples, which presents a simpler vision problem due to color differentiation, remains underdeveloped, the challenges in detecting and estimating the pose of similarly colored immature green fruits amidst canopy foliage are more complex. The study successfully utilized advanced machine vision and deep learning techniques, particularly the YOLO11 model and Visual Transformers, to address these challenges. The application of the YOLO11 model demonstrated high precision in both detection and pose estimation of immature fruitlets. Coupled with the depth insights provided by Visual Transformers, such as DPT and Depth Anything V2, the study enhanced the capabilities of automated thinning systems significantly. This integration not only improves the efficiency of orchard management but also reduces the physical strain and health risks associated with manual labor, thereby indicating a promising direction for enhancing precision agriculture practices, especially in handling immature fruitlets. This comprehensive experimental approach in commercial orchards provided performance comparisons between two widely used deep learning models (YOLO11 and YOLOv8) and investigated the efficacy of emerging visual transformers for cost-effective RGB to RGB-D mapping to validate fruitlet poses. The findings from these comparisons and validations lead to the following specific conclusions:
\begin{itemize}
    \item The YOLO11n model demonstrated exceptional performance, achieving the highest precision score of 0.91 among all tested configurations and recording one of the top mAP@50 values at 0.95, indicating its effectiveness in complex agricultural settings.
    \item In pose estimation tasks, YOLO11n matched its high detection precision with a pose estimation precision of 0.91. The YOLO11s variant excelled, achieving the highest mAP@50 at 0.94, which underscores its precision and robustness in accurately estimating poses.
    \item YOLO11n distinguished itself with superior processing speeds, achieving an inference time of only 2.7 milliseconds, substantially faster than other variants. This rapid processing capability highlights YOLO11n's suitability for real-time applications in agricultural automation, where efficiency is critical.
    
    \item Visual Transformers Evaluation: The evaluation of Visual Transformers revealed that Depth Anything V2 markedly outperformed the Dense Prediction Transformer in pose length validation. This was evidenced by its superior ability to convert low-cost RGB images into detailed RGB-D maps and generate accurate 3D point clouds. Initial measurements with Intel RealSense showed significant discrepancies, with a RMSE of 9.9845 and an MAE of 7.7444, highlighting substantial variations from ground truth data. However, a notable improvement was observed with the DPT model reducing the RMSE to 3.2935 and the MAE to 2.6237. Depth Anything V2 achieved the best results, demonstrating the lowest errors with a RMSE of 1.5261 and an MAE of 1.2809, showcasing its efficacy in enhancing precision in pose estimation of immature green fruitlets. 
\end{itemize}

\section{Future Work}
Building on the findings of this study, future research will aim to develop a robust motion planning system and an advanced robotic end effector, paving the way for a fully automated fruitlet thinning solution in commercial orchards. This vision system has demonstrated potential beyond just immature green apples, applicable to other tree fruits that resemble immature green apples during their pre-thinning stages, such as pears, plums, apricots, cherries, and peaches. The workflow validated here could be adapted to manage these fruits as well, enhancing the breadth and applicability of the technology. The successful deployment of a vision system capable of precise detection and pose estimation sets a solid foundation for future work, which will concentrate on developing and testing a working prototype in field conditions. This next phase will focus on integrating these technological advancements into a cohesive system that can operate efficiently within the dynamic environment of an orchard, aiming to revolutionize how fruit thinning is conducted, reducing labor costs and improving crop management.

\section{Acknowledgements and Funding }
This research is funded by the National Science Foundation and United States Department of Agriculture, National Institute of Food and Agriculture through the “AI Institute for Agriculture” Program (Award No.AWD003473). We extend our heartfelt gratitude to Zhichao Meng, Martin Churuvija, Astrid Wimmer, Randall Cason, Diego Lopez, Giulio Diracca, and Priyanka Upadhyaya for their invaluable efforts in data preparation and logistical support throughout this project. Special thanks to Dave Allan for granting orchard access. We also acknowledge the contribution of open-source platforms Roboflow (https://roboflow.com/), Ultralytics (https://docs.ultralytics.com/models/yolo11/), Hugging Face (https://huggingface.co/), and OpenAI (ChatGPT) for the models and implementation assistance in our project through their open-source platform.

\section{Author contributions statement}
R.S conceptualization, data curation, software, methodology, validation,  writing original draft. M.K editing and overall funding to supervisory  \cite{sapkota2025image, sapkota2025improved, sapkota2023creating}

\bibliographystyle{elsarticle-harv} 
\bibliography{example}






\end{document}